\documentclass[11pt,a4paper]{article}
\usepackage[hyperref]{emnlp-ijcnlp-2019}
\usepackage{times}
\usepackage{latexsym}
\usepackage{graphicx}
\usepackage{url}
\usepackage{breakurl}
\usepackage{bm}
\usepackage{multirow}
\usepackage{multicol}
\usepackage{mathrsfs}
\usepackage{amsmath}
\usepackage{amsfonts}

\aclfinalcopy 

\title{Knowledge Aware Conversation Generation with Explainable Reasoning over Augmented Graphs}

\author{
	Zhibin Liu,  Zheng-Yu Niu,  Hua Wu,   Haifeng Wang\\
	Baidu Inc., Beijing, China \\
	{\tt \{liuzhibin05,niuzhengyu,wu\_hua, wanghaifeng\}@baidu.com} \\
	}

\date{}

\begin{document}
\maketitle
\begin{abstract}
Two types of knowledge, triples from knowledge graphs and texts from documents, have been studied for knowledge aware open-domain conversation generation, in which graph paths can narrow down vertex candidates for knowledge selection decision, and texts can provide rich information for response generation. Fusion of a knowledge graph and texts might yield mutually reinforcing advantages, but there is less study on that. To address this challenge, we propose a knowledge aware chatting machine with three components, an \emph{augmented knowledge graph} with both triples and texts, knowledge selector, and knowledge aware response generator. For knowledge selection on the graph, we formulate it as a problem of \emph{multi-hop graph reasoning} to effectively capture conversation flow, which is more explainable and flexible in comparison with previous work. To fully leverage long text information that differentiates our graph from others, we improve a state of the art reasoning algorithm with \emph{machine reading comprehension technology}. We demonstrate the effectiveness of our system on two datasets in comparison with state-of-the-art models$\footnote{Data and codes are available at \url{https://github.com/PaddlePaddle/models/tree/develop/PaddleNLP/Research/EMNLP2019-AKGCM}\label{PaddleNLP}}$.
\end{abstract}

\section{Introduction}
One of the key goals of AI is to build a machine that can talk with humans when given an initial topic. To achieve this goal, the machine should be able to understand language with background knowledge, recall knowledge from memory or external resource, reason about these concepts together, and finally output appropriate and informative responses. Lots of research efforts have been devoted to chitchat oriented conversation generation \cite{Ritter2011,Shang2015}. 

However, these models tend to produce generic responses or incoherent responses for a given topic, since it is quite challenging to learn semantic interactions merely from dialogue data without help of background knowledge. Recently, some previous studies have been conducted to introduce external knowledge, either unstructured knowledge texts \cite{Ghazvininejad2018,Vougiouklis2016} or structured knowledge triples \cite{Liu2018,Young2018,Zhou2018} to help open-domain conversation generation by producing responses conditioned on selected knowledge. 

In the first research line, their knowledge graph can help narrowing down knowledge candidates for conversation generation with the use of prior information, e.g., triple attributes or graph paths. Moreover, these prior information can enhance generalization capability of knowledge selection models. But it suffers from information insufficiency for response generation since there is simply a single word or entity to facilitate generation. In the second line, their knowledge texts, e.g., comments about movies, can provide rich information for generation, but its unstructured representation scheme demands strong capability for models to perform knowledge selection or attention from the list of knowledge texts. Fusion of graph structure and knowledge texts might yield mutually reinforcing advantages for knowledge selection in dialogue systems, but there is less study on that. 

\begin{figure}[ht]
	\centering\includegraphics[width=3in,height=2.5in]{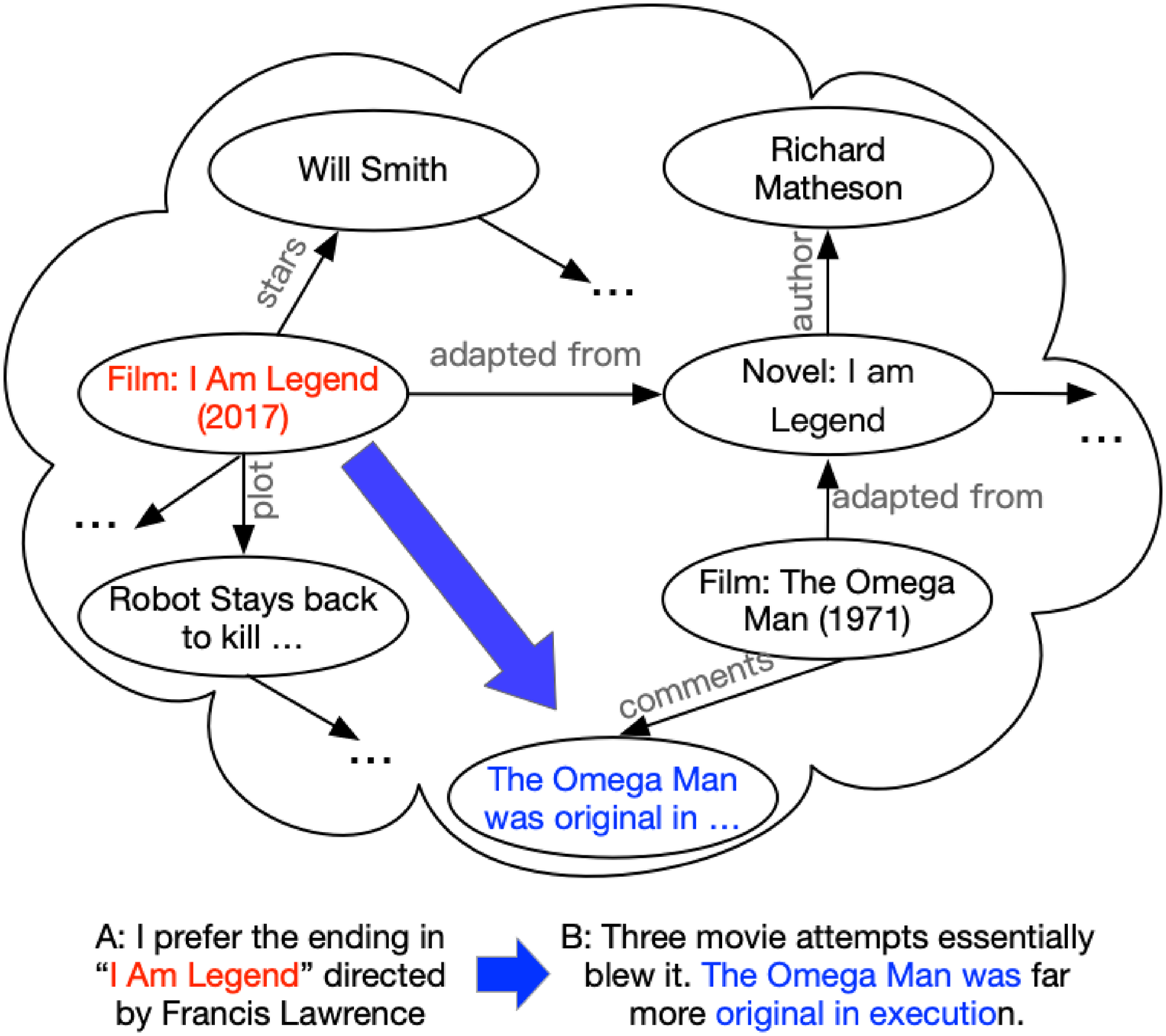}
	\caption{\footnotesize{Sample conversation with explainable reasoning, reflecting conversation flow, on an augmented knowledge graph.}}\label{fig:example}
\end{figure}

To bridge the gap between the two lines of studies mentioned above, we present an Augmented Knowledge Graph based open-domain Chatting Machine (denoted as \textbf{AKGCM}), which consists of knowledge selector and knowledge aware response generator. This two-stage architecture and graph based knowledge selection make our system to be explainable. Explainability is very important in e.g., information oriented chatting scenarios, where a user needs to know how new knowledge in chatbot's responses is linked to the knowledge in their utterances, or business scenarios, where a user cannot make a business decision without justification. 

To integrate texts into a knowledge graph, we take a factoid knowledge graph (KG) as its backbone, and align unstructured sentences of non-factoid knowledge with the factoid KG by linking entities from these sentences to vertices (containing entities) of the KG. Thus we augment the factoid KG with non-factoid knowledge, and retain its graph structure. Then we use this augmented KG to facilitate knowledge selection and response generation, as shown in Figure \ref{fig:example}. 

For knowledge selection on the graph, we adopt a deep reinforcement learning (RL) based reasoning model\cite{Das2018}, MINERVA, in which the reasoning procedure greatly reflects conversation flow as shown in Figure \ref{fig:example}. It is as robust as embedding based neural methods, and is as explainable as path based symbolic methods. Moreover, our graph differs from previous KGs in that: some vertices in ours contain long texts, not a single entity or word. To fully leverage this long text information, we improve the reasoning algorithm with machine reading comprehension (MRC) technology \cite{Seo2017} to conduct fine-grained semantic matching between an input message and candidate vertices.

Finally, for response generation, we use an encoder-decoder model to produce responses conditioned on selected knowledge.

In summary, we make following contributions:
\begin{itemize}
	\item This work is the first attempt that unifies knowledge triples and texts as a graph, and conducts flexible multi-hop knowledge graph reasoning in dialogue systems. Supported by such knowledge and knowledge selection method, our system can respond more appropriately and informatively. 
	\item Our two-stage architecture and graph based knowledge selection mechanism provide better model explainability, which is very important for some application scenarios.
	\item For knowledge selection, to fully leverage long texts in vertices, we integrate machine reading comprehension (MRC) technology into the graph reasoning process.
\end{itemize}


\section{Related Work}
\textbf{Conversation with Knowledge Graph:} There are growing interests in leveraging factoid knowledge \cite{Han2015,Liu2018,Zhu2017} or commonsense knowledge \cite{Young2018,Zhou2018} with graph based representation for generation of appropriate and informative responses. Compared with them, we augment previous KGs with knowledge texts and integrate more explainable and flexible multi-hop graph reasoning models into conversation systems. \citet{Wu2018} used document reasoning network for modeling of conversational contexts, but not for knowledge selection. 

\textbf{Conversation with Unstructured Texts:} With availability of a large amount of knowledge texts from Wikipedia or user generated content, some work focus on either modeling of conversation generation with unstructured texts \cite{Ghazvininejad2018,Vougiouklis2016,Xu2017}, or building benchmark dialogue data grounded on knowledge \cite{Dinan2019,Moghe2018}. In comparison with them, we adopt a graph based representation scheme for unstructured texts, which enables better explainability and generalization capability of our system.

\textbf{Knowledge Graph Reasoning:} Previous studies on KG reasoning can be categorized into three lines, path-based symbolic models \cite{Das2017,Lao2011}, embedding-based neural models \cite{Bordes2013,Wang2014}, and models in unifying
embedding and path-based technology \cite{Das2018,Lin2018,Xiong2017}, which can predict missing links for completion of KG. In this work, for knowledge selection on a graph, we follow the third line of works. Furthermore, our problem setting is different from theirs in that some of our vertices contain long texts, which motivates the use of machine reading technology for graph reasoning.

\textbf{Fusion of KG triples and texts:} In the task of QA, combination of a KG and a text corpus has been studied with a strategy of late fusion \cite{Gardner2017,Ryu2014} or early fusion \cite{Das2017b,Sun2018}, which can help address the issue of low coverage to answers in KG based models. In this work, we conduct this fusion for conversation generation, not QA, and our model can select sentences as answers, not restricted to entities in QA models.


\section{The Proposed Model}
\label{sec:model}
\subsection{Problem Definition and Model Overview}
\label{sec:3.1}
Our problem is formulated as follows: Let $\mathcal{G} = \{\mathcal{V},\mathcal{E}, \mathcal{L}^{\mathcal{E}}\}$ be an augmented KG, where $\mathcal{V}$ is a set of vertices, $\mathcal{E}$ is a set of edges, and $\mathcal{L}^{\mathcal{E}}$ is a set of edge labels (e.g., triple attributes, or vertex categories). Given a message $X =\{ x_{1}, x_2, ..., x_m\}$ and $\mathcal{G}$, the goal is to generate a proper response $Y = \{y_{1}, y_{2}, ..., y_{n}\}$ with supervised models. Essentially, the system consists of two stages: (1) knowledge selection: we select the vertex that maximizes following probability as an answer, which is from vertex candidates connected to $v_{X}$: 
\begin{equation}
v_{Y} = \arg\max_{v} P_{KS} (v |v_{X}, \mathcal{G}, X). 
\end{equation}
$v_{X}$ is one of vertices retrieved from $\mathcal{G}$ using the entity or words in $X$, and it is ranked as top-1 based on text similarity with $X$. Please see Equation \ref{eq:8} and \ref{eq:10} for computation of $P_{KS}(*)$; (2) response generation: it estimates the probability: 
\begin{equation}
P_{RG} (Y |X, v_{Y}) =  \prod_{t=1}^n P(y_{t}|y_{<t}, X, v_{Y}).
\end{equation}

\begin{figure}[t]
	\centering\includegraphics[width=3.1in]{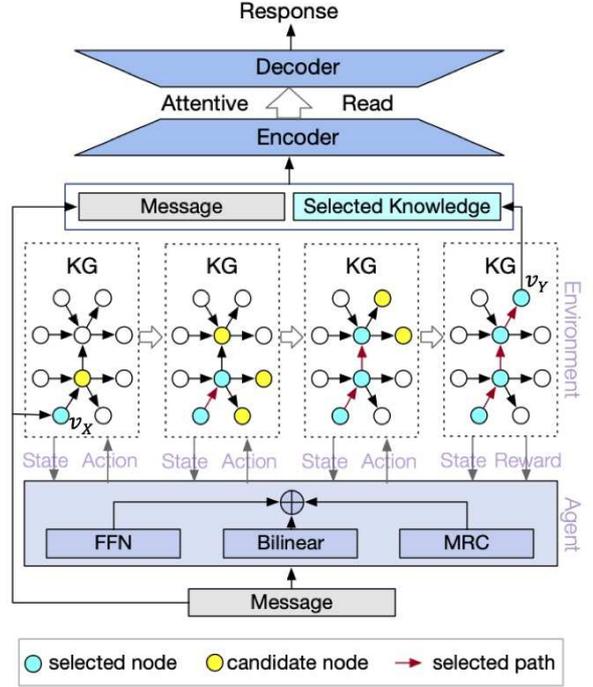}
	\caption{\footnotesize{The architecture of AKGCM.}}\label{fig:architecture}
\end{figure}

The overview of our Augmented Knowledge Graph based Chatting Machine (AKGCM) is shown in Figure \ref{fig:architecture}. The knowledge selector first takes as input a message $X =\{ x_{1}, x_2, ..., x_m\}$ and retrieves a starting vertex $v_{X}$ from $\mathcal{G}$ that is closely related to $X$, and then performs multi-hop graph reasoning on $\mathcal{G}$ and finally arrives at a vertex $v_{Y}$ that has the knowledge being appropriate for response generation. The knowledge aware response generator produces a response $Y = \{y_{1}, y_{2}, ..., y_{n}\}$ with knowledge from $v_{Y}$. At each decoding position, it attentively reads the selected knowledge text, and then generates a word in the vocabulary or copies a word in the knowledge text. 

For model training, each pair of [message, response] in training data is associated with ground-truth knowledge and its vertex ID (ground-truth vertex) in $\mathcal{G}$ for knowledge grounding. These vertex IDs will be used as ground-truth for training of knowledge selector, while the triples of [message, knowledge text, response] will be used for the training of knowledge aware generator. 

\subsection{Augmented Knowledge Graph}
\label{sec:akg}
Given a factoid KG and related documents containing non-factoid knowledge, we take the KG as a backbone, where each vertex contains a single entity or word, and each edge represents an attribute or a relation. Then we segment the documents into sentences and align each sentence with entries of the factoid KG by mapping entities from these sentences to entity vertices of the KG. Thus we augment the factoid KG with non-factoid knowledge, and retain its structured representation.


\subsection{Knowledge Selection on the Graph}
\textbf{Task Definition:}
We formulate knowledge selection on $\mathcal{G}$ as a finite horizon sequential decision making problem. It supports more flexible multi-hop walking on graphs, not restricted to one-hop walking as done in previous work \cite{Han2015,Zhou2018,Zhu2017}. 

As shown in Figure \ref{fig:architecture}, we begin by representing the environment as a deterministic partially observed Markov decision process (POMDP) on $\mathcal{G}$ built in Section \ref{sec:akg}. Our RL based agent is given an input query of the form $(v_{X},X)$. Starting from vertex $v_{X}$ corresponding to $X$ in $\mathcal{G}$, the agent follows a path in the graph, and stops at a vertex that it predicts as the answer $v_Y$. Using a training set of known answer vertices for message-response pairs, we train the agent using policy gradients \cite{Williams1992} with control variates. 

The difference between the setting of our problem and previous KG reasoning lies in that: (1) the content of our input queries is not limited to entities and attributes; (2) some vertices in our graph contains long texts, while vertices in previous KGs just contain a single entity or short text. It motivates us to make a few improvements on previous models, as shown in Equation (\ref{eq:5}), (\ref{eq:6}), and (\ref{eq:7}). 

Next we elaborate the 5-tuple $(\mathcal{S},\mathcal{O},\mathcal{A},\delta,\mathcal{R})$ of the environment, and policy network.

\textbf{States:} A state $S_t \in \mathcal{S}$ at time step $t$ is represented by $S_t = (v_{t},v_{X},X,v_{gt})$ and the state space consists of all valid combinations in $\mathcal{V}\times\mathcal{V}\times\mathcal{X}\times\mathcal{V}$, where $v_{t}$ is current location of the RL agent, $v_{gt}$ is the ground-truth vertex, and $\mathcal{X}$ is the set of all possible $X$.

\textbf{Observations:} The complete state of the environment cannot be observed. Intuitively, the agent knows its current location $(v_{t})$ and $(v_{X},X)$, but not the ground-truth one $(v_{gt})$, which remains hidden. Formally, the observation function $\mathcal{O}:\mathcal{S}\overrightarrow{}\mathcal{V}\times\mathcal{V}\times\mathcal{X}$ is defined as $\mathcal{O}(S_t = (v_{t},v_{X},X,v_{gt})) = (v_{t},v_{X},X)$.

\textbf{Actions:} The set of possible actions $\mathcal{A}_{S_t}$ from a state $S_t$ consists of all outgoing edges of the vertex $v_t$ in $\mathcal{G}$. Formally $\mathcal{A}_{S_t} = \{(v_t , l_e, v_d) \in \mathcal{E} : S_t = (v_{t},v_{X},X,v_{gt}), l_e \in \mathcal{L}^{\mathcal{E}}, v_d \in \mathcal{V} \} \cup  \{(S_t, \emptyset, S_t)\}$. It means an agent at each state has option to select which outgoing edge it wishes to take with the label of the edge $l_e$ and destination vertex $v_d$. We limit the length of the action sequence (horizon length) up to a fixed number (e.g., $T$) of time steps. Moreover, we augment each vertex with a special action called ‘NO\_OP’ which goes from a vertex to itself. This decision allows the agent to remain at a vertex for any number of time steps. It is especially helpful when the agent has managed to reach a correct vertex at a time step $t < T$ and can continue to stay at the vertex for the rest of the time steps. 

\textbf{Transition:} The environment evolves deterministically by just updating the state to the new vertex according to the edge selected by the agent. Formally, the transition function $\delta:\mathcal{S}\times\mathcal{A}\overrightarrow{}\mathcal{S}$ is defined by $\delta(S_t,A) = (v_d,v_{X},X,v_{gt})$, where $S_t=(v_{t},v_{X},X,v_{gt})$ and $A=(v_t , l_e, v_d)$. $l_e$ is the label of an edge connecting $v_t$ and $v_d$, and $v_d$ is destination vertex.

\textbf{Rewards:} After $T$ time steps, if the current vertex is the ground-truth one, then the agent receives a reward of $+1$ otherwise $0$. Formally, $\mathcal{R}(S_T)=I\{v_{T} =v_{gt}\}$, where $S_T=(v_{T},v_{X},X,v_{gt})$ is the final state.

\textbf{Policy Network:} We design a randomized non-stationary policy $\bm{\pi} = (\bm{d_0,d_1,...,d_{T-1}})$, where $d_t=P(\mathcal{A}_{S_t}) $ is a policy at time step $t$. In this work, for each $d_t$, we employ \emph{a policy network with three components} to make the decision of choosing an action from all available actions ($\mathcal{A}_{S_t}$) conditioned on $X$.

The first component is a history dependent \emph{feed-forward network} (FFN) based model proposed in \cite{Das2018}. We first employ a LSTM to encode the history $H_t = (H_{t-1},A_{t-1},O_t)$ as a continuous vector $\bm{h_t} \in \mathbb{R}^{2d}$, where $H_t$ is the sequence of observations and actions taken. It is defined by:
\begin{equation}
\bm{h_t} = LSTM(\bm{h_{t-1}},[\bm{a_{t-1}};\bm{o_t}]),
\end{equation}
where $\bm{a_{t-1}}$ is the embedding of the relation corresponding to the label of the edge the agent chose at time $t-1$ and $\bm{o_t}$ is the embedding of the vertex corresponding to the agent's state at time $t$.

Recall that each possible action represents an outgoing edge with information of the edge relation label $l_e$ and destination vertex $v_d$. So let $[\bm{l_e};\bm{v_d}]$ denote an embedding for each action $A \in \mathcal{A}_{S_t}$, and we obtain the matrix $\bm{A_t}$ by stacking embeddings for all the outgoing edges. Then we build a two-layer feed-forward network with ReLU nonlinearity which takes in the current history representation $h_t$ and the representation of $X$ ($\bm{e^{new}_X}$). We use another single-layer \emph{feed-forward network} for computation of $\bm{e^{new}_X}$, which accepts the original sentence embedding of $X$ ($\bm{e_X}$) as input. The updated FFN model for action decision is defined by:
\begin{equation}
\begin{split}
P_{FFN}(\mathcal{A}_{S_t})= \bm{A_t}(\bm{W_2}ReLU\\(\bm{W_1} [\bm{h_t};\bm{o_t};\bm{e^{new}_X}] + \bm{b_1}) + \bm{b_2}),
\end{split}
\end{equation}
\begin{equation}
\label{eq:5}
\bm{e^{new}_X}=ReLU(\bm{W_Xe_X} + \bm{b_X}).
\end{equation}

Recall that in our graph, some vertices contain long texts, differentiating our graph from others in previous work. The original reasoning model \cite{Das2018}, MINERVA, cannot effectively exploit the long text information within vertices since it just learns embedding representation for the whole vertex, without detailed analysis of text in vertices. To fully leverage the long text information in vertices, we employ two models, \emph{a machine reading comprehension model} (MRC) \cite{Seo2017} and \emph{a bilinear model}, to score each possible $v_d$ from both global and local view. 

For scoring from global view, (1) we build a document by collecting sentences from all possible $v_d$, (2) we employ the MRC model to predict an answer span ($span_{aw}$) from the document, (3) we score each $v_d$ by calculating a ROUGE-L score vector of $v_d$'s sentence with $span_{aw}$ as the reference, shown as follows: 
\begin{equation}
\label{eq:6}
P_{MRC}(\mathcal{A}_{S_t}) = ROUGE(Text(V_d), span_{aw}).
\end{equation}
Here, $Text(\cdot)$ represents operation of getting text contents, and $ROUGE(\cdot)$ represents operation of calculating ROUGE-L score. We see that the MRC model can help to determine which $v_d$ is the best based on global information from the whole document.  

For scoring from local view, we use another \emph{bilinear model} to calculate similarity between $X$ and $v_d$, shown as follows:
\begin{equation}
\label{eq:7}
P_{Bi}(\mathcal{A}_{S_t}) = \bm{V_dW_{B}e_{X}}.
\end{equation}

Finally, we calculate a sum of outputs of the three above-mentioned models and outputs a probability distribution over the possible actions from which a discrete action is sampled, defined by:
\begin{equation}
\label{eq:8}
\begin{split}
P(\mathcal{A}_{S_t})=softmax(\alpha P_{FFN}(\mathcal{A}_{S_t})+\\ \beta P_{Bi}(\mathcal{A}_{S_t})
+\gamma P_{MRC}(\mathcal{A}_{S_t})),
\end{split}
\end{equation}
\begin{equation}
A_t \sim Sample(P(\mathcal{A}_{S_t})),
\end{equation}
\begin{equation}
\label{eq:10}
P_{KS} (v_{d} |v_{X}, \mathcal{G}, X)=P(\mathcal{A}_{S_{T-1}}).
\end{equation}
Please see Section \ref{sec:3.1} for definition of $P_{KS}(*)$. When the agent finally arrives at $S_T$, we obtain $v_T$ as the answer $v_Y$ for response generation.

\textbf{Training:}
For the policy network ($\pi_\theta$) described above, we want to find parameters $\theta$ that maximize the expected reward:
\begin{equation}
\begin{split}
J(\theta)=\mathbb{E}_{(v_0,X,v_{gt})\sim D}\mathbb{E}_{A_0,...,A_{T-1}\sim\pi_\theta}\\
[\mathcal{R}(S_T)|S_0 =(v_0,v_0,X,v_{gt})],
\end{split}
\end{equation}
where we assume there is a true underlying distribution $D$, and $(v_0,X,v_{gt}) \sim D$.

\subsection{Knowledge Aware Generation}
Following the work of \citet{Moghe2018}, we modify a text summarization model \cite{See2017} to suit this generation task. 

In the summarization task, its input is a document and its output is a summary, but in our case the input is a [selected knowledge, message] pair and the output is a response. Therefore we introduce two RNNs: one is for computing the representation of the selected knowledge, and the other for the message. The decoder accepts the two representations and its own internal state representation as input, and then compute (1) a probability score which indicates whether the next word should be generated or copied, (2) a probability distribution over the vocabulary if the next word needs to be generated, and (3) a probability distribution over the input words if the next word needs to be copied. These three probability distributions are then combined, resulting in $P(y_{t}|y_{<t}, X, v_{Y})$, to produce the next word in the response. 

\section{Experiments and Results}
\subsection{Datasets}
We adopt two knowledge grounded multi-turn dialogue datasets for experiments, shown as follows: 

\textbf{EMNLP dialog dataset \cite{Moghe2018}} This Reddit dataset contains movie chats from two participants, wherein each response is explicitly generated by copying or modifying sentences from background knowledge such as IMDB's facts/plots, or Reddit's comments about movies. We follow their data split for training, validation and test\footnote{We use the single-reference mixed-short test set for evaluation. Please see their paper for more details.}. Their statistics can be seen in Table \ref{table:statistics}. 

\textbf{ICLR dialog dataset \cite{Dinan2019}} This wizard-of-wiki dataset contains multi-turn conversations from two participants. One participant selects a beginning topic, and during the conversation the topic is allowed to naturally change. The two participants are not symmetric: one will play the role of a knowledgeable expert while the other is a curious learner. We filter their training data and test data by removing instances without the use of knowledge and finally keep 30\% instances\footnote{Their ground-truth responses should have high ROUGE-L scores with corresponding ground-truth knowledge texts.} for our study since we focus on knowledge selection and knowledge aware generation. Their statistics can be seen in Table \ref{table:statistics}. For models (Seq2Seq, HRED) without the use of knowledge, we keep the original training data for them.

\begin{table}[t]  
	\centering
	\begin{tabular}{ l l l l  }
		\hline
		\multicolumn{4}{ c }{EMNLP dialog dataset}   \\
		\hline
		\multicolumn{2}{ l }{Conversational Pairs}  &\multicolumn{2}{l }{Augmented KG}  \\
		\hline
		\#Train. pairs& 34486  & \#Vertices & 117373  \\
		\#Valid. pairs& 4388  & \#Relations & 11 \\
		\#Test pairs& 4318  & \#Triples & 251138  \\
		\hline
		\hline
		\multicolumn{2}{ l }{Factoid know.}  &\multicolumn{2}{l }{Non-f. know.}  \\
		\hline
		\#Total Ver. &  21028& \#Total Ver. & 96345 \\
		\#Used Ver.  & 2620 & \#Used Ver. & 27586  \\
		\hline
		\hline
		\multicolumn{4}{ c }{ICLR dialog dataset} \\
		\hline
		\multicolumn{2}{ l }{Conversational Pairs}  &\multicolumn{2}{l }{Augmented KG}  \\
		\hline
		\#Train. pairs& 19484  & \#Vertices & 290075  \\
		\#Valid. pairs& 1042  & \#Relations & 5 \\
		\#Test pairs& 1043  & \#Triples & 2570227  \\
		\hline
		\hline
		\multicolumn{2}{ l }{Factoid know.}  &\multicolumn{2}{l }{Non-f. know.}  \\
		\hline
		\#Total Ver. & NA & \#Total Ver. & 290075 \\
		\#Used Ver.  & NA & \#Used Ver. & 10393  \\
		\hline
	\end{tabular}
	\caption{The upper/lower two tables show statistics of the EMNLP/ICLR datasets and corresponding augmented knowledg  graphs. \#Used Ver. means total num of vertices that used for response generation. For more details, please visit our data sharing URL.}\label{table:statistics}
\end{table}

\subsection{Experiment Settings}
We follow the existing work to conduct both automatic evaluation and human evaluation for our system. We also compare our system with a set of carefully selected baselines, shown as follows. 

\textbf{Seq2Seq:} We implement a sequence-to-sequence model (Seq2Seq) \cite{Sutskever2014}, which is widely used in open-domain conversational systems.

\textbf{HRED:} We implement a hierarchical recurrent encoder-decoder model \cite{Serban2016}. 

\textbf{MemNet:} We implement an end-to-end knowledge-MemNet based conversation model \cite{Ghazvininejad2018}. 


\textbf{GTTP:} It is an end-to-end text summarization model \cite{See2017} studied on the EMNLP data. We use the code$\footnote{\url{https://github.com/nikitacs16/Holl-E}\label{emnlp_holle}}$ released by \citet{Moghe2018}, where they modify GTTP to suit knowledge aware conversation generation.

\textbf{BiDAF+G:} It is a Bi-directional Attention Flow based QA Model (BiDAF) \cite{Seo2017} that performs best on the EMNLP dataset. We use the code\textsuperscript{\ref{emnlp_holle}} released by \citet{Moghe2018}, where they use it to find the answer span from a knowledge document, taking the input message as the query. Moreover, we use a response generator (as same as ours) for NLG with the predicted knowledge span.

\textbf{TMemNet:} It is a two-stage transformer-MemNet based conversation system that performs best on the ICLR dataset \cite{Dinan2019}. We use the code$\footnote{\url{https://parl.ai/projects/wizard_of_wikipedia/}}$ released by the original authors.

\textbf{CCM:} It is a state-of-the-art knowledge graph based conversation model \cite{Zhou2018}. We use the code$\footnote{\url{https://github.com/tuxchow/ccm}}$ released by the original authors and then modify our graph to suit their setting by selecting each content word from long text as an individual vertex to replace our long-text vertices.

\textbf{AKGCM:} It is our two-stage system presented in Section \ref{sec:model}. We implement our knowledge selection model based on the code$\footnote{\url{https://github.com/shehzaadzd/MINERVA}}$ by\cite{Das2018} and that\textsuperscript{\ref{emnlp_holle}} by \cite{Moghe2018}. We use BiDAF as the MRC module, shown in Equation (\ref{eq:6}), and we train the MRC module on the same training set for our knowledge selection model. We implement the knowledge aware generation model based on the code of GTTP\textsuperscript{\ref{emnlp_holle}} released by \cite{Moghe2018}. We also implement a variant AKGCM-5, in which top five knowledge texts are used for generation, and other setting are not changed.

\begin{table*}[t] 
	\centering
	\begin{tabular}{ p{1.8cm} | r r r r |r r r r}
		\hline
		  &\multicolumn{4}{|c|}{EMNLP dialog dataset} &\multicolumn{4}{|c}{ICLR dialog dataset}\\
		&\small{BLEU-4}&\small{ROUGE-2}& \small{ROUGE-L}& \small{Hit@1}&\small{BLEU-4}&\small{ROUGE-2}& \small{ROUGE-L}& \small{Hit@1}\\
		Model&\scriptsize{ (\%)}  & \scriptsize{ (\%)} & \scriptsize{ (\%)}& \scriptsize{ (\%)}&\scriptsize{ (\%)}  & \scriptsize{ (\%)} & \scriptsize{ (\%)} & \scriptsize{ (\%)}\\
		\hline
		Seq2seq& 1.59 & 5.73 & 14.49 & NA     & 0.17 & 1.01 & 7.02 & NA \\
		HRED& 2.08 & 8.83 & 18.13 & NA        & 0.23 &1.08 & 7.32 & NA\\
		MemNet &  5.86 & 10.64 & 18.48 & NA    & 0.89 & 2.33 & 11.84& NA\\
		GTTP & 11.05 & 17.70 &   25.13 & NA    & 6.74 & 7.18	 & \textbf{17.11} & NA\\
		CCM & 2.40 & 4.84 & 17.70 & NA         & 0.86 & 1.68 & 12.74 & NA \\
		BiDAF+G & \textbf{32.45} & \textbf{31.28} & \textbf{36.95} & 40.80      & 6.48 & 6.54 & 15.56 & 17.40\\
		TMemNet & 8.92 & 13.15 & 19.97 & 38.10      & 1.09 & 1.86 & 8.51 & 16.80 \\		
		AKGCM-5& 13.29 &13.12 & 21.22 & \textbf{42.04}      & \textbf{6.94} & \textbf{7.38} & 17.02 & \textbf{18.24}\\
		AKGCM& 30.84 & 29.29 & 34.72 & \textbf{42.04}      & 5.52 & 6.10 & 15.46 & \textbf{18.24}\\
		\hline
	\end{tabular}
	\caption{Results of automatic evaluations on the two datasets.}\label{table:auto_eval}
\end{table*}

\newcommand{\tabincell}[2]{\begin{tabular}{@{}#1@{}}#2\end{tabular}}  
\begin{table*}[t] 
	\centering 
	\begin{tabular}{ p{1.6cm} | c c | c c}
		\hline
		&\multicolumn{2}{|c|}{EMNLP dialog dataset} &\multicolumn{2}{|c}{ICLR dialog dataset}\\
		&  Appr.  &  Infor. &  Appr.  &  Infor.  \\ 
		&  * vs. AKGCM & * vs. AKGCM & * vs. AKGCM-5 & * vs. AKGCM-5 \\
		Model & Win/Tie/Lose & Win/Tie/Lose & Win/Tie/Lose & Win/Tie/Lose \\
		\hline
		Seq2seq& 0.04/0.42/0.54 & 0.05/0.21/0.74    & 0.00/0.10/0.90 & 0.00/0.11/0.89     \\
		HRED& 0.03/0.50/0.47 & 0.03/0.27/0.70     & 0.01/0.14/0.85 & 0.01/0.14/0.85  \\
		MemNet & 0.03/0.43/0.54 & 0.03/0.23/0.74    & 0.00/0.19/0.81 & 0.00/0.17/0.83   \\
		GTTP & 0.03/0.52/0.45 & 0.10/0.42/0.48   & 0.07/0.73/0.20 & 0.12/0.68/0.20   \\
		CCM & 0.01/0.18/0.81 & 0.01/0.15/0.84   & 0.00/0.17/0.83 & 0.00/0.16/0.84   \\
		BiDAF+G & 0.04/0.83/0.13 & 0.07/0.79/0.14  & 0.04/0.61/0.35 & 0.04/0.56/0.40    \\
		TMemNet & 0.04/0.50/0.46 & 0.05/0.36/0.59  & 0.01/0.25/0.74 & 0.01/0.21/0.78 \\
		\hline
	\end{tabular}
	\caption{Results of human evaluations on the two datasets. AKGCM (or AKGCM-5) outperforms all the baselines significantly (sign test, p-value $<$ 0.05) in terms of the two metrics.}\label{table:human_eval}
\end{table*}

\subsection{Automatic Evaluations}
\textbf{Metrics:} Following the work of \cite{Moghe2018}, we adopt BLEU-4 \cite{Papineni2002}, ROUGE-2 \cite{Lin2004a} and ROUGE-L \cite{Lin2004b} to evaluate how similar the output response is to the reference text. We use Hit@1 (the top 1 accuracy) to evaluate the performance of knowledge selection.

\textbf{Results:} As shown in Table \ref{table:auto_eval}, AKGCM (or AKGCM-5) can obtain the highest score on test set in terms of Hit@1, and the second highest scores in terms of BLEU-4, ROUGE-2 and ROUGE-L, surpassing other models, except BiDAF, by a large margin. It indicates that AKGCM has a capability of knowledge selection better than BiDAF and TMemNet, and generates more informative and grammatical responses. We notice that from EMNLP dataset to ICLR dataset, there is a significant performance drop for almost all the models. It is probably due to that the quality of ICLR dataset is worse than that of EMNLP dataset. A common phenomenon of ICLR dataset is that the knowledge used in responses is loosely relevant to input messages, which increases the difficulty of model learning.

\subsection{Human Evaluations}
\textbf{Metrics:} We resort to a web crowdsourcing service for human evaluations. We randomly sample 200 messages from test set and run each model to generate responses, and then we conduct pair-wise comparison between the response by AKGCM and the one by a baseline for the same message. In total, we have 1400 pairs on each dataset since there are seven baselines. For each pair, we ask five evaluators to give a preference between the two responses, in terms of the following two metrics: (1) appropriateness (Appr.), e.g., whether the response is appropriate in relevance, and logic, (2) informativeness (Infor.), whether the response provides new information and knowledge in addition to the input message, instead of generic responses such as “This movie is amazing”. Tie is allowed. Notice that system identifiers are masked during evaluation. 

\textbf{Annotation Statistics:} We calculate the agreements to measure inter-evaluator consistency. For appropriateness, the percentage of test instances that at least 2 evaluators give the same label (2/3 agreement) is 98\%, and that for at least 3/3 agreement is 51\%. For informativeness, the percentage for at least 2/3 agreement is 98\% and that for at least 3/3 agreement is 55.5\%.

\textbf{Results:} In Table \ref{table:human_eval}, each score for win/tie/lose is the percentage of messages for which AKGCM (or AKGCM-5) can generate better/almost same/worse responses, in comparison with a baseline. We see that our model outperforms all the baselines significantly (sign test, p-value $<$ 0.05) in terms of the two metrics on the two datasets. Furthermore, our model can beat the strongest baseline, BiDAF. It demonstrates the effectiveness of our graph reasoning mechanism that can use global graph structure information and exploit long text information. Our data analysis shows that both Seq2Seq and HRED tend to generate safe responses starting with ``my favorite character is'' or ``I think it is''. Both Memnet and TMemnet can generate informative responses. But the knowledge in their responses tends to be incorrect, which is a serious problem for knowledge aware conversation generation. Our results show that GTTP and BiDAF are very strong baselines. It indicates that the attention mechanism (from machine reading) for knowledge selection and the copy mechanism can bring benefits for knowledge aware conversation generation. Although CCM have the mechanisms mentioned above, this model is good at dealing with structured triples rather than long texts. It may explain the inferior performance of CCM in our problem setting.  

We find that many responses are likely to be a simple copy of the selected knowledge, which is a reflection of the characteristics of the datasets. In the two datasets, some sentences in the background knowledge are directly used as responses to the messages. Therefore, the NLG module is likely to copy content from the selected knowledge as much as possible for generation. Moreover, the summarization model GTTP tends to copy words from its input message as its output due to its generation mechanism.
Table \ref{table:two_dataset_better} presents the examples in which AKGCM (AKGCM-5) performs better than other models on two dataset.

\subsection{Model Analysis}

\begin{table}[t]
	\centering 
	\begin{tabular}{p{2.5cm} r  r  r }
		\hline
		& \small{BLEU-4}    &\small{ROUGE-2} &\small{ROUGE-L}  \\
		Model variant &\scriptsize{ (\%)}  & \scriptsize{ (\%)} & \scriptsize{ (\%)} \\ 
		\hline
		w/o non-factoid knowledge  & 0.61 &  0.84 & 1.52  \\
		
		\hline
		w/o Bilinear + MRC   & 15.12 & 14.15 & 21.09  \\
		w/o MRC  & 18.41 & 17.79 & 24.59  \\
		w/o Bilinear  & 28.62 & 28.19 & 32.84  \\
		\hline
		Full model & 30.84 & 29.29 & 34.72 \\
		\hline
	\end{tabular}
	\caption{Results of ablation study for AKGCM on EMNLP dataset. We also include results of the full model for comparison. }\label{table:Ablation}
\end{table}

\begin{figure}[ht]
	\centering\includegraphics[width=3.1in]{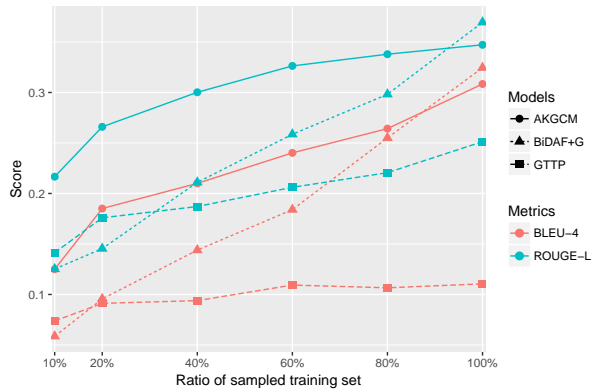}
	\caption{\footnotesize{Results of how each model performs on EMNLP dataset when we gradually reduce the ratio of sampled training data from 100\% to 10\% (from right to left).}}\label{fig:MG}
\end{figure}

\begin{table*}[t]
	\centering
	\small
	\begin{tabular}{|p{1.5cm}|p{6.0cm}| p{6.0cm}|}
		\hline
		Dataset & EMNLP dataset (Movie Name: Monsters University) & ICLR dataset (Topic Name: Green) \\
		\hline
		Message &  I love comic scene in movies though it makes me want to see it over again  & How is green color produced?  \\
		\hline
		Ground-truth response & I will be definitely be seeing it a second and possibly third time .   & Green is evoked by light, a dominant wavelength of about 495570 nm. it is between blue and yellow.  \\
		\hline
		Seq2seq & I think it 's just plain funny  & Yellow is the color between green and orange on the spectrum of visible light  \\
		\hline
		HRED & A stone cold classic !  & There are different types of UNK there are different types of UNK  \\
		\hline
		MemNet & I thought it was a great movie, but it is a great film.  & It lies between violet and green on the spectrum of light.  \\
		\hline
		GTTP & I think this movie was worth it , for i will be definitely be seeing it a second and possibly third time. & Yes , green is the color between blue and yellow on the visible spectrum . \\
		\hline
		CCM & I think it was a worth \_UNK , but I still enjoyed it intense than the fraternities . & The energy is chemical on the Several minerals and tradition . \\
		\hline
		BiDAF+G & It not only serves as a great climax , but something that we can once again commend Pixar for telling great stories in out of the box thinking that we should think paying for the \$ 10 dollar ticket is worth it . & It 's the color between blue and the color yellow . \\
		
		Selected knowledge & It not only serves as a great climax, but something that we can once again commend Pixar for telling great stories in out of the box thinking that we should think paying for the \$10 dollar ticket is worth it. And boy did I think this movie was worth it, ...  & Green is the color between blue and yellow on the visible spectrum. \\
		
		\hline
		TMemNet & Man and i think this movie was worth it , i will definitely be seeing if it was a second time . & It ' s been around since the middle ages , it was associated with royalty . \\
		
		
		\hline
		AKGCM (AKGCM-5) & And boy did I think this movie was worth it , for I will be definitely be seeing it a second and possibly third time . & Green is evoked by light which has a dominant wavelength of roughly 495570 nm . \\
		
		Selected knowledge & And boy did I think this movie was worth it, for I will be definitely be seeing it a second and possibly third time. &  ... It is evoked by light which has a dominant wavelength of roughly 495570 nm. Several minerals have a green color, including the emerald, ... \\
		
		\hline
		
	\end{tabular}
	\caption{Examples in which AKGCM performs better than other models on two dataset.}\label{table:two_dataset_better}
\end{table*}

\textbf{AKGCM without (w/o) Non-factoid Knowledge:} To verify contribution of non-factoid knowledge, we remove non-factoid knowledge from augmented KG in test procedure, and report the performance of our system with only factoid knowledge in Table \ref{table:Ablation}. We see that without non-factoid knowledge from EMNLP dataset, the performance of our system drops significantly in terms of BLEU and ROUGE. It indicates that non-factoid knowledge is essential for knowledge aware conversation generation.

\textbf{AKGCM w/o the MRC Model or Bilinear One:} For ablation study, we implement a few variants of our system without the bilinear model or MRC for knowledge selection. Results of these variants are reported in Table \ref{table:Ablation}. If we compare the performance of our full model with its variants, we find that both MRC and the bilinear model can bring performance improvement to our system. It indicates that the full interaction between messages and knowledge texts by their attention mechanism is effective to knowledge selection.

\textbf{Model Generalization:} As shown in Figure \ref{fig:MG}, we gradually reduce the size of training data, and then AKGCM can still manage to achieve acceptable performance, even when given extremely small training data (around 3,400 u-r pairs at the x-axis point of 10\%). But the performance of the strongest baseline, BIDAF+G, drops more dramatically in comparison with AKGCM. It indicates that our graph reasoning mechanism can effectively use the graph structure information for knowledge selection, resulting in better generalization capability of AKGCM.

\textbf{Model Explainability:} We check the graph paths traversed by our system for knowledge selection and try to interpret what heuristics have been learned. We find that our system can learn to visit different types of vertices conditioned on conversational contexts, e.g.. selecting comment vertices as responses for utterances starting with ``what do you think''. These results suggest that AKGCM may also be a good assistive tool for discovering new algorithms, especially in cases when the graph reasoning are new and less well-studied for conversation modeling.

\section{Conclusion} 
In this paper, we propose to augment a knowledge graph with texts and integrate it into an open-domain chatting machine with both graph reasoning based knowledge selector and knowledge aware response generator. Experiments demonstrate the effectiveness of our system on two datasets compared to state-of-the-art approaches. 

\section*{Acknowledgments}
We would like to thank the reviewers for their insightful comments. This work was supported by the Natural Science Foundation of China (No.61533018).

\newpage
\bibliography{emnlp-ijcnlp-2019}
\bibliographystyle{acl_natbib}


\end{document}